\documentclass{article}

 \PassOptionsToPackage{numbers, compress}{natbib}

\usepackage[preprint]{nips_2018}
\usepackage{graphicx}
\usepackage{subcaption}
\usepackage{mathtools}
\usepackage{array}
\usepackage{layouts}

\usepackage[utf8]{inputenc} 
\usepackage[T1]{fontenc}    
\usepackage{hyperref}       
\usepackage{url}            
\usepackage{booktabs}       
\usepackage{amsfonts}       
\usepackage{nicefrac}       
\usepackage{microtype}      
\newcommand{\etal}{\textit{et al.}}

\title{DSCnet: Replicating Lidar Point Clouds \\ with Deep Sensor Cloning}

\author{
  Paden Tomasello*, Sammy Sidhu, Anting Shen, Matthew W. Moskewicz, \\
  {\bf Nobie Redmon, Gayatri Joshi, Romi Phadte, Paras Jain, Forrest Iandola} \\
  DeepScale, Inc. \\
  Mountain View, CA, USA \\
  *\texttt{paden@deepscale.ai}
}

\begin{document}

\maketitle

\begin{abstract}

Convolutional neural networks (CNNs) have become increasingly popular for solving a variety of computer vision tasks, ranging from image classification to image segmentation.
Recently, autonomous vehicles have created a demand for depth information, which is often obtained using hardware sensors such as Light detection and ranging (LIDAR).
Although it can provide precise distance measurements, most LIDARs are still far too expensive to sell in mass-produced consumer vehicles, which has motivated methods to generate depth information from commodity automotive sensors like cameras.

In this paper, we propose an approach called Deep Sensor Cloning (DSC).
The idea is to use Convolutional Neural Networks in conjunction with inexpensive sensors to replicate the 3D point-clouds that are created by expensive LIDARs.
To accomplish this, we develop a new dataset (DSDepth) and a new family of CNN architectures (DSCnets).
While previous tasks such as KITTI depth prediction use an interpolated RGB-D images as ground-truth for training, we instead use DSCnets to directly predict LIDAR point-clouds. 
When we compare the output of our models to a \$75,000 LIDAR, we find that our most accurate DSCnet achieves a relative error of 5.77\% using a single camera and 4.69\% using stereo cameras.

\end{abstract}

\section{Introduction and Related Work}

\subsection{Introduction}

Convolutional neural networks have become quintessential for solving a variety of computer vision tasks such as classification \cite{15}, semantic segmentation \cite{16}, and depth prediction \cite{1}.  
Most previous attempts of using CNNs for depth prediction attempt to predict depth for each pixel of the image, and the ground truth comes from either a virtual dataset as ShapeNet \cite{ShapeNet}, or a dataset that interpolates missing values from a LIDAR as in the KITTI dataset \cite{10}.
In this paper, we propose a method called Deep Sensor Cloning, which directly regresses on the output of LIDAR using a variety of sensor inputs.

\subsection{Motivation} \label{sec:motivation}

The promise of autonomous vehicles has inspired numerous advancements in both perception systems and automotive sensors.
Perception systems create three-dimensional representation of their environment, so autonomous vehicles can safely and effectively navigate and control themselves.
Any effective perception system must provide accurate depth information, which has traditionally been obtained using a sensor like radar or LIDAR.
Radars, which are already in mass-produced vehicles, only provide coarse depth measurements for objects like cars or motorcycles.
While this may be sufficient for adaptive cruise control or emergency braking, full autonomy will require more fine grained depth information.
LIDAR provides perception systems with precise distance measurements of its environment in the form of a three-dimensional point cloud.
Given these advantages, LIDAR (particularly Velodyne HDL-32 and HDL-64) is now used in prototype autonomous vehicles developed by Uber, Zoox, and Cruise Automation.

While accomplishing the task of depth prediction well these Velodyne LIDARs are very expensive, which means that LIDAR is now a barrier of entry for autonomous vehicles to the mass market, because of their cost, power consumption, and complexity.
For example, Velodyne's HDL-64 LIDAR, a model similar to the one used in DARPA Grand Challenge, consumes 60W of power and is estimated to cost \$75,000, limiting their use to vehicles in robo-taxi services.
Velodyne's cheapest model the VLP-16 contains $\frac{1}{4}$ of the number lasers as the HDL-64, and is still estimated to cost \$8,000, a price still too expensive for all but the most expensive consumer vehicles.
When trading notes with others who have done field work in autonomous driving, we have heard criticism that the Velodyne VLP-16 and HDL-64 LIDARs rely on precisely calibrated internal components, which can make them prone to be less accurate or break from standard wear and tear.

More and more companies are developing LIDAR products, and a comparison of some  currently-available LIDAR devices is shown in Table \ref{tab:lidar-comparison}
\footnote{LIDAR information was obtained through \url{http://www.woodsidecap.com/wp-content/uploads/2018/04/Yole\_WCP-LiDAR-Report\_April-2018-FINAL.pdf}}.
As can be seen in this table, there is a tradeoff between between price and resolution.
The Ibeo Scala LIDAR has a price-point low enough for certain mass-produced vehicles, but its resolution is not high enough to support advanced autonomous capabilities.
In Section \ref{sec:model-architecture}, we present a method to fuse data from a low-cost, low-resolution LIDAR with camera images, to clone a high-cost, high-resolution LIDAR at a fraction of its price.

\begin{table}[h]
  \centering
  \begin{tabular}{|p{2cm}|p{2.5cm}|p{1cm}|p{2cm}|p{1.5cm}|p{1.5cm}|}
    \hline
    Manufacturer & Model & Number of Lasers & Data Rate (pts/sec) & Power (Watts) & Cost (USD) \\ [0.5ex]
    \hline\hline
    Velodyne & VLS-128 & 128 & 9,600,000  & Unknown & Unknown \\
    \hline
    Velodyne & HDL-64 & 64 & 1,300,000  & 60 & \$75,000 \\
    \hline
    Velodyne & HDL-32 & 32 & 700,000 & 12 & \$30,000 \\
    \hline
    Velodyne & VLP-16  & 16 & 300,000 & 8 & \$8,000 \\
    \hline
    Robosense  & RS-LIDAR-32  & 32 & 640,000  & 13.5 & \$16,800 \\
    \hline
    Ouster  & OS-1 & 64 & 1,310,720  & Unknown & \$12,000 \\
    \hline
    Ibeo  & Lux  & 4-8 & Unknown & 7-10 & \$10,000-20,000 \\
    \hline
    Ibeo  & Scala  & 4  & Unknown  & 7 & \$600 \\
    \hline
  \end{tabular}
  \caption{\label{tab:lidar-comparison} Comparison of some of the LIDAR sensors that can be purchased today. In most cases, the number of lasers is the vertical resolution; for example, a 4-laser LIDAR has just 4 pixels of vertical resolution. And, the Data Rate is equivalent to $framerate*resolution$. \\ {\tiny Thanks to Woodside Capital Partners for this table.} }
\end{table}

There are numerous companies attempting to improve the LIDAR hardware, including Luminar, Quanergy, and Innoviz, which are not shown in Table~\ref{tab:lidar-comparison} because (a) they have not announced their sensor's specifications, and/or (b) because their sensors are not available for purchase yet.
In the future, some of these attempts may produce a durable and high-resolution LIDAR that is cheap enough for the mass market.
At the moment however, there does not appear to be a solution which provides the detail needed for an advanced perception system, at a cost needed for mass-market production.
As the industry waits for further development, deep learning has recently provided alternative methods to predict the depth from various cheaper sensors.

\subsection{Related Work}

Depth estimation creates a dense depth map, or an RGB-D image, given no explicit input information information about depth.
From the 1980s until around 2014, the most widely discussed (and probably most widely used) approach for depth estimation from cameras was {\em stereo-matching}~\cite{LucasKanadeStereo}~\cite{3}.
Stereo-matching identifies point-correspondences across two cameras and then uses relative position of the two cameras to reconstruct the depth of each point in the image.
However, the principal weakness of stereo-matching is the robustness of the point-correspondence algorithms.
While better feature engineering (e.g. the invention of SIFT features in 1999~\cite{SIFT,SIFTstereo}) has led to incremental improvements in the accuracy of point-correspondence algorithms, stereo algorithms still frequently fail to find point-correspondences in numerous situations, such as featureless walls, vegetation, scenes with significant scale changes, and so on~\cite{HowardStereoOdometry}.
In 2014, Eigen \etal published one of the first in a series of papers on using convolutional neural networks to directly regress depth from an single input image, dismissing notions that stereo disparity is essential for depth prediction ~\cite{1}.
Since 2014, further innovations in CNN architecture and loss-functions have yielded additional improvements in depth estimation \cite{4, 5, 6}.

A key challenge for depth estimation tasks is collecting the training and evaluation data. 
Two popular datasets used for depth estimation are KITTI Depth and Make3D, which provide synchronized camera images and dense depth-maps that are derived from interpolating sparse LIDAR point-clouds~\cite{1, 18}.
ShapeNet is a dataset that uses simulation to generate 3D imagery, with ground-truth 3D information stored in voxels~\cite{ShapeNet}.
We compare and contrast these datasets in Table~\ref{tab:depth-dataset}.
Notably, none of these allow researchers to train a CNN to clone a real sensor; rather, each of these datasets provides ground-truth based on (a) sensor data that is postprocessed with interpolation that hallucinates that data that doesn't exist, or (b) simulation of an imaginary sensor.
In contrast to this, our dataset presented in Section~\ref{sec:DSDepth} enables CNNs to be trained to directly clone a \$75,000 LIDAR sensor.

\begin{table}[h]
	\centering
	\footnotesize
	\begin{tabular}{|p{2cm}|p{1.3cm}|p{1.5cm}|p{1.1cm}|p{1.6cm}|p{1.8cm}|p{1.5cm}|}
		\hline
		& Inference Data & \multicolumn{4}{|c|}{Ground Truth (GT) Training Data} & \\
		\hline
		Name & Inference Sensors & GT Sensors & GT Data Type & GT \mbox{Coordinate} System & GT Uses \mbox{Interpolation?} & Number of samples \\ [0.5ex]
		\hline\hline
		Kitti Depth~\cite{10} & Cameras & Velodyne HDL-64 LIDAR & Depth Map & Cartesian & Yes & 93,000 \\
		\hline
		Make3D~\cite{18} & Mono Camera & custom \mbox{LIDAR} & Depth Map & Cartesian & Yes & 400  \\
		\hline
		ShapeNet~\cite{ShapeNet} & Simulation & Simulation & Voxels & Cartesian & No & 53,000 \\
		\hline
		DSDepth (ours)  & Cameras and Scala & HDL-64 & PCDM & Polar & No & 78,968 \\
		\hline
	\end{tabular}
	\caption{\label{tab:depth-dataset} Comparison of depth datasets. DSDepth will be explained in Section~\ref{sec:DSDepth}.}
\end{table}

\subsection{Key Contributions}

Unlike the previous approaches, we present a solution called Deep Sensor Cloning (DSC), which regresses the depth output of the LIDAR directly. In the process of developing and evaluating DSC, we have made the following key contributions:

\begin{enumerate}
  \item A novel method of regressing depth using a Point Cloud as a Dense Matrix (PCDM) output format
  \item A template for leveraging sensor fusion in CNNs
  \item A new set of metrics for evaluating depth estimation in the context of autonomous driving
\end{enumerate}

The rest of this paper is organized as follows.
In Section~\ref{sec:approach}, we present our approach to collecting and representing multi-sensor data, and we introduce the DSCnet family of CNN architectures.
Next, Section~\ref{sec:train-eval-methodology} describes our approach for training and evaluating our CNNs.
Then, in Section~\ref{sec:results}, we present qualitative and quantitative results on how well our CNNs can predict depth using only low-cost sensors.
We conclude in Section~\ref{sec:conclusion}.

\section{Approach} \label{sec:approach}

In this section, we explain our approach for using deep neural networks to ingest data from inexpensive sensors and output a point-cloud that is similar to what is produced by a \$75,000 Velodyne LIDAR. 
To do this, we develop a custom data format (called a PCDM), create a new dataset (called DSDepth), design a new family of CNNs (called DSCnets), and propose a loss function (called Sparse L2 Loss).
We devote the rest of this section to explaining these concepts.

\subsection{PCDM Data Format} \label{sec:approach-pcdm}

Traditional datasets utilizing LIDAR such as KITTI \citep{10}, store LIDAR point-cloud data in a cartesian coordinate system, where each point in the scan is represent as a triplet of (x, y, z). 
This format can be difficult to use in neural networks, since it is both sparse and three dimensional. 
VoxelNet uses this format as an input to a 3D object detector by dividing the point cloud in 3D voxels and then transforming each voxel using a voxel feature encoding \citep{12}.
While this showed promising results using a sparse input to a CNN, sparse outputs present new and different challenges. To overcome to challenges, We created a novel format for storing point-cloud we call Point Cloud as a Dense Matrix, or PCDM. 

\begin{figure}[h]
  \begin{subfigure}[c]{0.5\textwidth}
    \includegraphics[width=\textwidth]{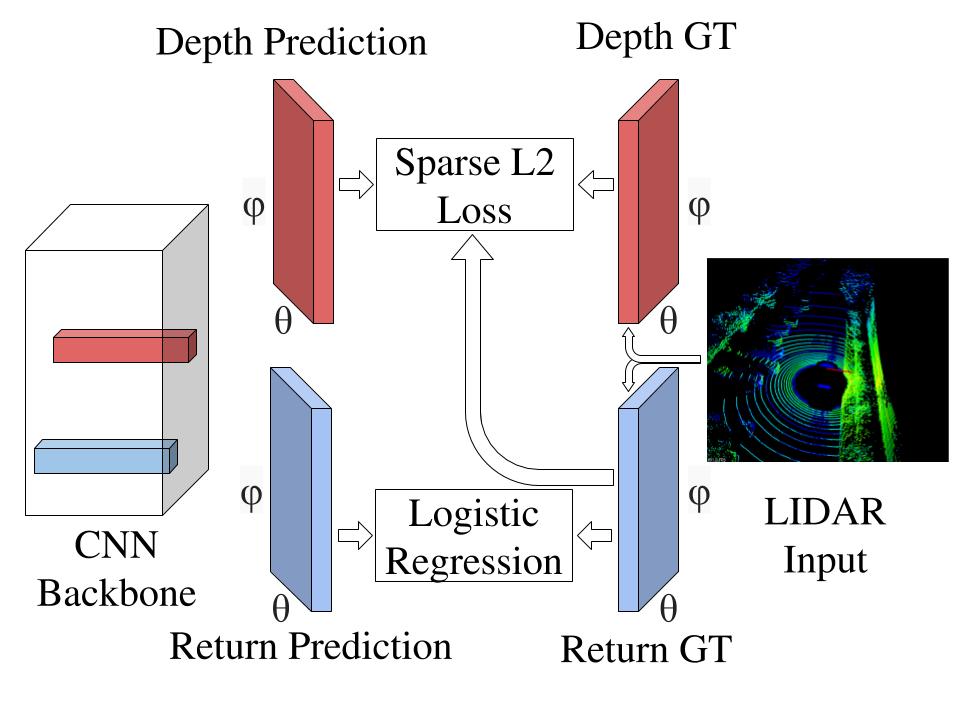}
    \caption{Training using the PCDM format.}
    \label{fig:pcdm_training}
  \end{subfigure}
  \unskip\ \vrule\ \vrule\
  \begin{subfigure}[c]{0.5\textwidth}
    \includegraphics[width=\textwidth]{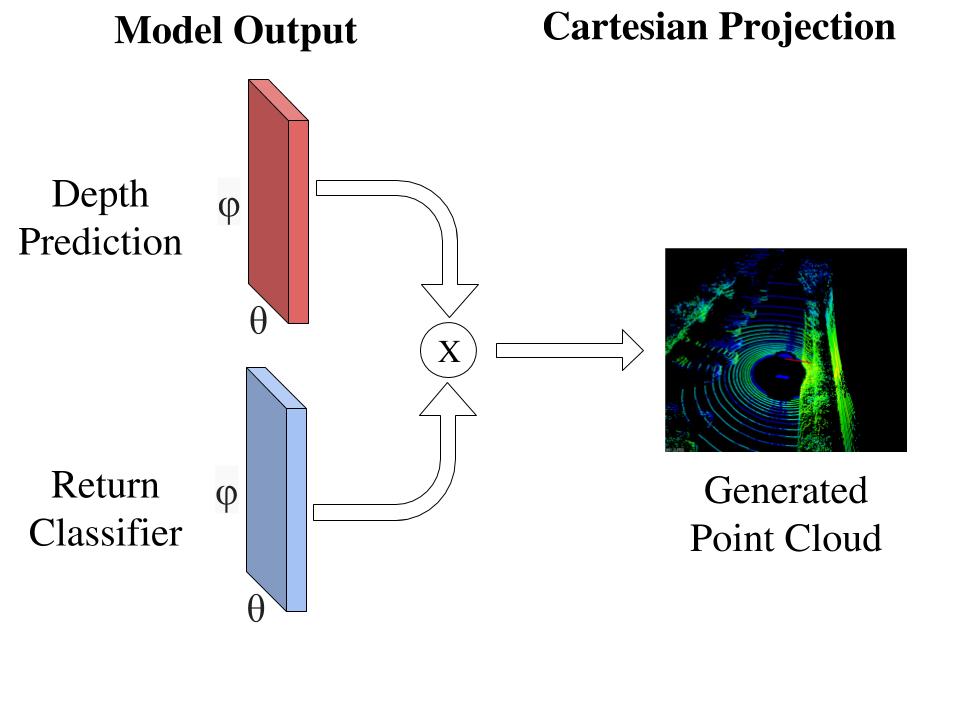}
    \caption{Point-cloud reconstruction.}
    \label{fig:pcdm_output_format}
  \end{subfigure}
  \caption{Training and inference methodology using the PCDM format.}
  \label{fig:pcdm_format}
\end{figure}

The PCDM format is composed of a "Depth" matrix and a "Return" matrix.
In the "Depth" matrix, each column corresponds to an angular position of the LIDAR, and each row corresponds to a azimuth positioning of a laser, and the value at each point is the measured distance from the LIDAR.

A LIDAR will not usually have a distance measurement for all locations in its scan, because an object may be too far away, or not have enough reflectance for the LIDAR to receive any reflected signal.
For any of scan that does not have a return, the LIDAR will mark its distance as 0.
If we simply tried to regress on the sparse "Depth" matrix, CNNs would have difficulty learning the difference between a small distance measurement and a non-return scan.
In order to encode this additional information, we create a separate "Return" matrix, which stores a binary value representing whether the LIDAR had a return or not.
As can be scene in Figure \ref{fig:pcdm_training}, when we train our CNN, we mask the gradients for any pixels that do not have a LIDAR return, so the network can ignore any non-returns.

In addition to using the "Return" ground truth to block the gradient, we also create a classifier head for a network to try to replicate its values.
By doing so, we can create realistic LIDAR point clouds by masking our depth prediction by our "Return" classifier, as can be seen in Figure \ref{fig:pcdm_output_format}.

\subsection{DSDepth Dataset} \label{sec:DSDepth}

Now that we have defined data format, we describe how we collected a new dataset called DSDepth.
In designing this dataset, our goal is to accurately represent two sets of hardware:
\begin{itemize}
  \item {\bf Expensive Sensors}: Hardware that can be deployed on an expensive (but small) group of cars that are used for data-collection and R\&D.
  \item {\bf Inexpensive Sensors}: Hardware that can be deployed on millions of reasonably low-cost cars that are used every day by consumers and fleet-operators.
\end{itemize}

\begin{figure}[h]
	\begin{subfigure}[l]{0.4\textwidth}
		\includegraphics[width=\textwidth]{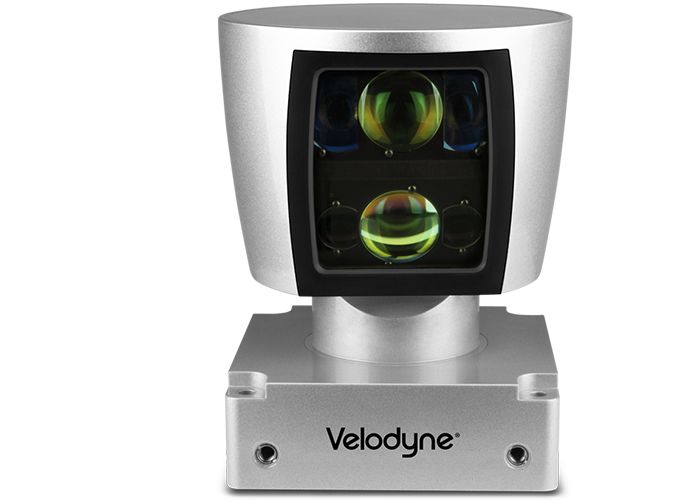}
		\caption{Velodyne HDL-64 LIDAR (\$75,000). This is part of our "Expensive" sensor set.}
		\label{fig:HDL-64}
	\end{subfigure}
	\hfill
	\begin{subfigure}[r]{0.4\textwidth}
		\includegraphics[width=\textwidth]{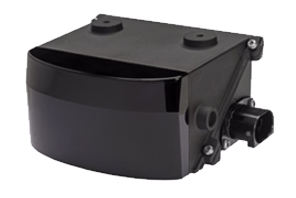}
		\caption{Ibeo Scala LIDAR (under \$1000). \\ This is part of our "Inexpensive" sensor set.}
		\label{fig:scala}
	\end{subfigure}
	\caption{LIDAR sensors mounted on DeepScale's data collection vehicle. Note that the Velodyne has 16x more vertical resolution than the Scala.}
	\label{fig:lidars}
\end{figure}

In DSDepth, the sole "expensive sensor" is a Velodyne HDL-64 LIDAR (Figure~\ref{fig:HDL-64}). 
With a price of \$75,000, the HDL-64 is the one of the most expensive LIDARs in Table~\ref{tab:lidar-comparison}.
While autonomous R\&D vehicles often have other sensors such as cameras and radars, the LIDAR is often the go-to sensor for depth sensing. 
Given that our goal in this paper is to produce depth estimates (in the form of point-clouds), we think it is reasonable to use the HDL-64 as the sole "expensive sensor" for the purposes of this paper.

\begin{figure}[h]
	\begin{subfigure}[l]{0.5\textwidth}
		\includegraphics[width=\textwidth]{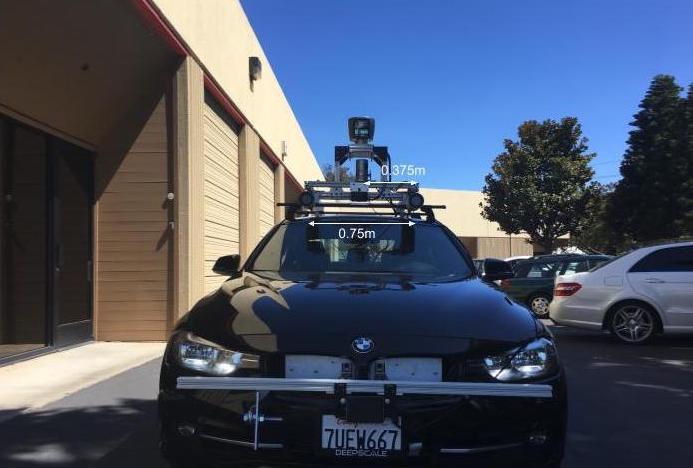}
		\caption{Front view}
		\label{fig:dsf_car_front}
	\end{subfigure}
	\begin{subfigure}[r]{0.5\textwidth}
		\includegraphics[width=\textwidth]{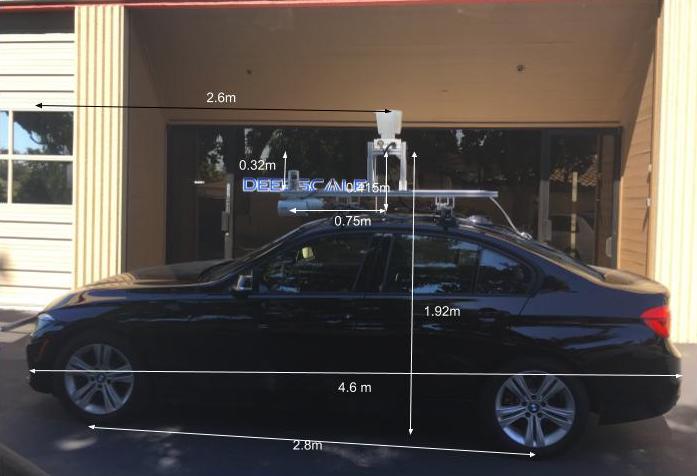}
		\caption{Side view}
		\label{fig:dsf_car_side}
	\end{subfigure}
	\caption{DeepScale's data collection vehicle.}
	\label{fig:dsf_car}
\end{figure}

In DSDepth, we have also have a set of three "inexpensive sensors."
Two of these sensors are cameras, which we have mounted side-by-side on the roof of the car. 
The third "inexpensive" sensor is an Ibeo Scala LIDAR (Figure~\ref{fig:scala}).
While the \$75,000 Velodyne HDL-64 is too expensive to be deployed on mass-produced cars, the sub-\$1000 Ibeo Scala LIDAR has been deployed on mass-produced vehicles such as the Audi A8.\footnote{\url{https://www.businesswire.com/news/home/20180705005220/en/Global-In-vehicle-LiDAR-Industry-Outlook-2022-Expected}}
Note that, while the Scala and HDL-64 are both LIDARs, the Scala has $\frac{1}{16}$ the vertical resolution of the HDL-64, and the Scala's vertical field of view is almost an order-of-magnitude narrower than the HDL-64's vertical field of view.\footnote{The HDL-64 as a 26.9-degree vertical field-of-view, and the Scala has only a 3.2-degree vertical field-of-view.}
In Figure~\ref{fig:dsf_car}, we show how these and other sensors are integrated onto our data-collection car.

\subsubsection{Data Capture Implementation}

We synchronized our sensors using a triggering system, which captures an image when the Velodyne LIDAR was pointed toward the front of the vehicle.
We then collected the Velodyne LIDAR data samples from the previous 180 degrees, and next 180 degrees, and we transformed them into a PCDM using the strategy mentioned in Section \ref{sec:approach-pcdm}.
In our training procedure, we crop the PCDM so that all points are included in the field of field of our input images.
Further, we capture the most-recent full scan from the inexpensive Scala LIDAR.

\subsection{DSCnet Model Architectures} \label{sec:model-architecture}

In contrast to traditional stereo vision algorithms, we propose a family of Deep Sensor Cloning models (called {\em DSCnets}), which do not require any information of the camera intrinsics or extrinsics for registering with the LIDAR.
Rather, we allow the model to learn how to best leverage multiple sensors using end-to-end training.
We additionally designed a sensor fusion template that allows us to quickly experiment with various sensor configurations.

\begin{figure}[h]
  \begin{subfigure}[l]{0.45\textwidth}
    \includegraphics[width=\textwidth]{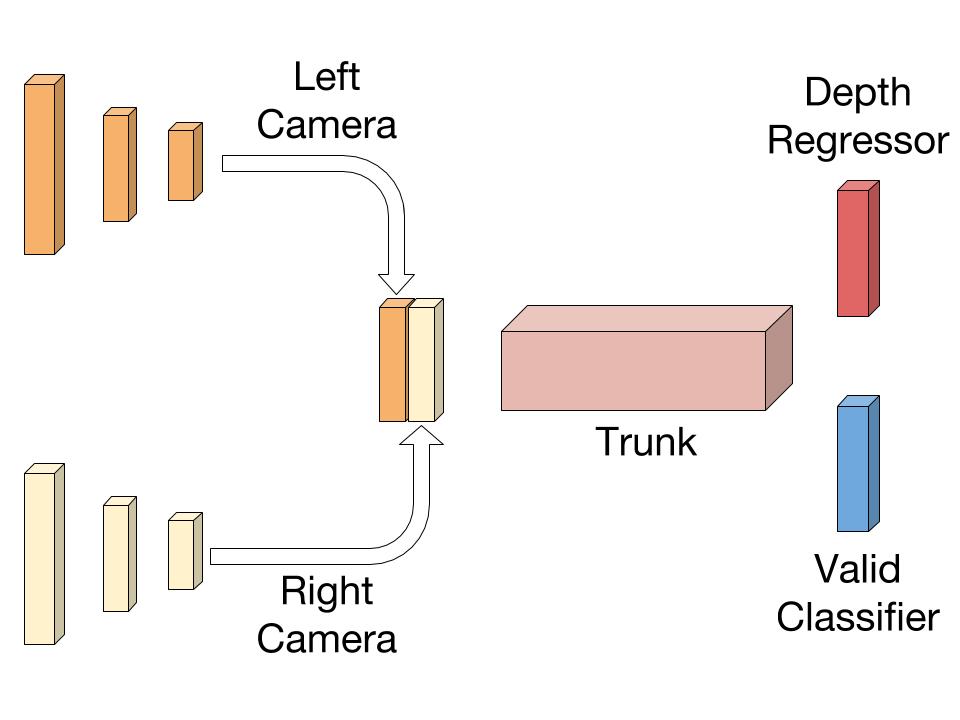}
    \caption{Stereo Camera Model}
    \label{fig:stereo_model}
  \end{subfigure}
  \unskip\ \vrule\ \vrule\
  \begin{subfigure}[r]{0.45\textwidth}
    \includegraphics[width=\textwidth]{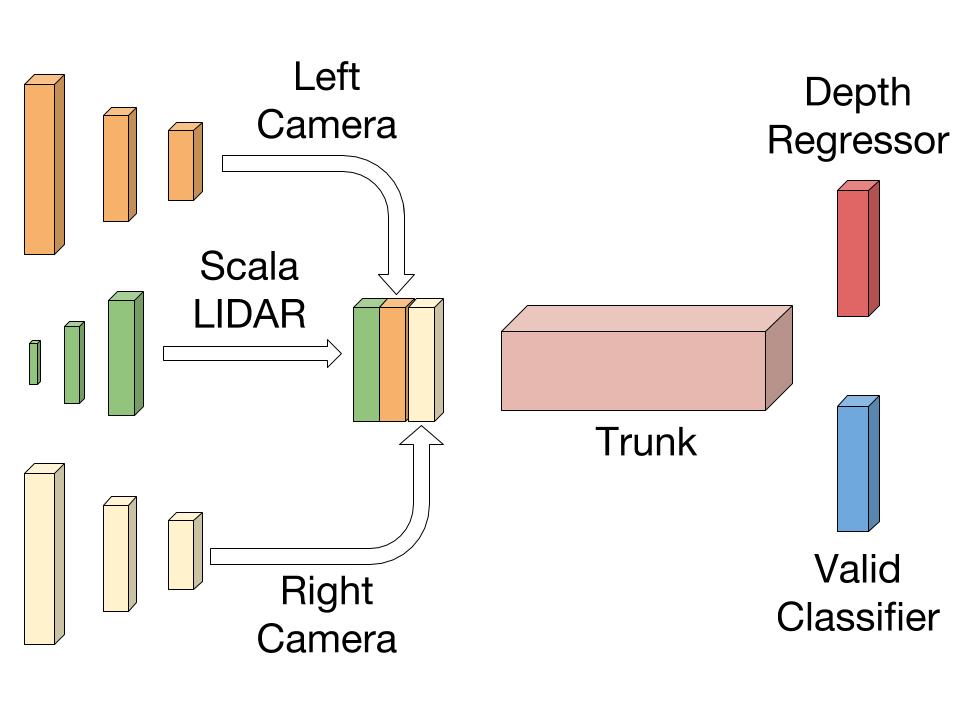}
    \caption{Stereo + Scala LIDAR Model}
    \label{fig:stereo-scala-model}
  \end{subfigure}
  \caption{Two examples of our DSCnet architectures}
  \label{fig:sensor-fusion-architectures}
\end{figure}

One of our goals in designing the DSCnet family of CNNs was to enable ourselves to quickly experiment with various input sensor sets, and we accomplished that as follows.
For each sensor, we created an independent branch of convolutions or deconvolution to both resize the data and learn features that are relevant to that particular sensor. 
Downstream of these sensor-specific networks, we add a "Trunk", which assumes an input of equal size to the Velodyne LIDAR PCDM, 64 by 256 in our experiments.
For our experiements, our "Trunk" is a  V-net architecture (inspired by \cite{14}).
Figure \ref{fig:stereo_model} shows an example of a model which fuses images from a left and right camera, and Table \ref{tab:image-resize-branch} shows the layer parameters of the camera's "resize" branch.

\begin{table}[h]
  \centering
  \begin{tabular}{c c c c c c}
    \hline
    \# of Units & Channels & Kernel Dimensions & Stride & Input Size & Output Size \\
    \hline
    \hline
    2 & 8 & (3,3) & (1,1) & (576, 768) & (576, 768) \\
    \hline
    1 & 16 & (5,5) & (3,3) & (576, 768) & (192, 256) \\
    \hline
    2 & 16 & (3,3) & (1,1) & (192, 256) & (192, 256) \\
    \hline
    1 & 32 & (5,5) & (3,1) & (192, 256) & (64, 256) \\
    \hline
    2 & 32 & (3,3) & (1,1) & (64, 768) & (64, 256) \\
    \hline
    \hline
  \end{tabular}
  \caption{\label{tab:image-resize-branch} Image Resize Branch of DSCnet}
\end{table}

The benefit of this approach is that adding additional sensors only requires creating a new branch into the concatenation operator of the network.
Additionally, it's much easier to compare sensor configuration, because the trunk network backbone is held constant.
An example architecture utilizing the Scala 4 Beam LIDAR is shown in Figure \ref{fig:stereo-scala-model}, and Table \ref{tab:scala-resize-branch} shows the layer parameters of Scala "resize" branch.

As dicussed in Section \ref{sec:motivation}, the Scala LIDAR is a much cheaper but lower resolution LIDAR compared to Velodyne HDL-64.
Using this model, we can create a point-cloud with the same resolution as the HDL-64, but at a fraction of its cost.

\begin{table}[h]
  \centering
  \begin{tabular}{c c c c c c}
    \hline
    \# of Units & Channels & Kernel Dimensions & Stride & Input Size & Output Size \\
    \hline
    \hline
    2 & 64 & (3,3) & (1,1) & (4, 192) & (4, 192) \\
    \hline
    1 & 128 & (3,3) & (1,3) & (4, 192) & (4, 64) \\
    \hline
    2 & 128 & (3,3) & (1,1) & (4, 64) & (4, 64) \\
    \hline
    1 & 64 & (3,3) & $(\frac{1}{2}, \frac{1}{2})$  & (4, 64) & (8, 128) \\
    \hline
    2 & 64 & (3,3) & (1,1) & (8, 128) & (8, 128) \\
    \hline
    1 & 32 & (3,3) & $(\frac{1}{2}, \frac{1}{2})$  & (8, 128) & (16, 256) \\
    \hline
    2 & 32 & (3,3) & (1,1) & (16, 256) & (16, 256) \\
    \hline
    1 & 16 & (3,3) & $(\frac{1}{2}, 1)$  & (16, 256) & (32, 256) \\
    \hline
    2 & 16 & (3,3) & (1,1) & (32, 256) & (32, 256) \\
    \hline
    1 & 16 & (3,3) & $(\frac{1}{2}, 1)$  & (32, 256) &  (64, 256)\\
    \hline
    2 & 16 & (3,3) & (1,1) & (64, 256) & (64, 256) \\
    \hline
    \hline
  \end{tabular}
  \caption{\label{tab:scala-resize-branch} Scala LIDAR Resize Branch of DSCnet}
\end{table}

\subsection{Sparse L2 Loss} \label{sec:loss}

When training DSCnet, we use separate loss functions for (a) regressing the distance measurements and (b) classifying the valid return in the PCDM.
For the classifier, we use a logistic regression loss.
As mentioned above, we do not backprop the gradient for scanned points that have no return, so we utilize a modified Least Squares Error, which we call Sparse L2 Loss.
It is defined as:

$$ L = \frac{1}{N}\sum_{i=0}^n ((D_i - f(X_i)*V_i) $$

where $Di$ is Depth Ground Truth, $f(X_i)$ is depth prediction, and $V_i$ is the "Return" mask, as described in section \ref{sec:approach-pcdm}.

Finally, we block the gradient from the classifier one layer before the loss function, so the trunk network is only trained to predict distance. Empirically, we found this to be sufficient to train the classifier head.

\section{Training and Evaluation Methodology} \label{sec:train-eval-methodology}

\subsection{Training Routine}

Our DSCnet models was trained on 58853 training samples and evaluated on 20115 validation samples.
We use a stochastic gradient descent optimizer with a learning rate of 0.013, momentum of 0.9, and weight decay of 0.0005, and decrease the learning rate by a factor of 0.2 every 60,000 iterations. We use a batch size of 48 and train across 3 Nvidia Titan Xp GPUs. 

\subsection{Evaluation Metrics} \label{sec:eval-metrics}

A number of metrics have been established in the research community to evaluate the correctness of depth-estimation algorithms.
These metrics include:
\begin{enumerate}
  \item Relative absolute error (percent): $ \frac{1}{N} \sum_{y} \frac{|y - y* |}{y*} $
  \item Relative squared error (percent): $ \frac{1}{N} \sum_{y} \frac{(y - y*)^2}{y*^2} $
  \item Root mean squared error of inverse depth [1/km]: $  \sqrt{ \frac{1}{N}\sum_{y} || \frac{1}{y} - \frac{1}{y*} ||^2  } $
  \item Scale invariant logarithmic error [1/km]: $ \frac{1}{N} \sum_{i} d_i^2 - \frac{1}{N^2} \left( \sum_i d_i \right)^2  $ where $ d_i = \log{y_i} - \log{y_i*} $
\end{enumerate}

where $ y $ is the predicted distance, and $ y* $ is the distance ground truth.

In Table~\ref{tab:kitti-leaderboard}, we show a snapshot of the current state-of-the-art results on the leaderboard for the KITTI Depth challenge~\cite{1}.
Out of the metrics shown on Table~\ref{tab:kitti-leaderboard}, we find Absolute Relative Error (absErrorRel) to be particularly intuitive. 
When driving a car, when we encounter an object that is 1 meter away, we care far more about 1-meter error than we do for an object that is 100 meters away.
The absErrorRel metric takes this into account -- a 1 meter error on a 100-meter-away object is worth the same absErrorRel penalty as a 0.01-meter error on a 1-meter-away object.

\begin{table}[h]
	\centering
	\footnotesize
	\begin{tabular}{|p{1cm}|p{1cm}|p{1cm}|p{1.3cm}|p{1.2cm}|p{.7cm}|p{.8cm}|p{1cm}|p{1cm}|p{1.1cm}|}
		\hline
		Dataset & CNN & Model Input &  absErrorRel & sqErrorRel & iRMSE & SILog & Return Classifier Error & GFLOP & Parameter Size (MB) \\ [0.5ex]
		\hline\hline
		KITTI Depth & DORN~\cite{21} & Mono camera & 8.78 & 2.23 & 12.98 & 11.77 & N/A & N/A & N/A \\ 
		\hline
	\end{tabular}
	\caption{\label{tab:kitti-leaderboard} Snapshot of the top result on the KITTI Depth leaderboard as of November 2018~\cite{1}.}
\end{table}

We have also added a few new metrics to our evaluation that are not included in the KITTI leaderboard.
In particular, since we are training DSCnet to mimic the sparsity pattern of an expensive Velodyne LIDAR, we report the accuracy of our return-classifier (see Section~\ref{sec:loss}).
Further, for our experiments in the next section, we will report the model's resource utilization in terms of computation (GFLOP per inference) and parameter file size (in megabytes).

\subsection{Metric Zones} \label{sec:metric-zones}

In autonomous driving applications, some areas are more critical than others to have accurate depth information.
In adaptive cruise control for example, the distance measurements directly in front of the vehicle are much more important than those to the side.
In order to create a better evaluation our models, we designed metric zones for a few different autonomous vehicle applications, and calculated the above metrics for each of these zones.
The metric zones are defined in Table \ref{tab:metric-zones}.

\begin{center}
\begin{table}[h]
 \begin{tabular}{|p{4cm}|p{2.5cm}|p{2.5cm}|p{2.5cm}|}
 \hline
   Name & Min Distance (m)  & Max Distance (m) & Horizontal Field of View (degrees)  \\ [0.5ex]
 \hline\hline
 Parking Assist & 0 & 10 & 44 \\
 \hline
   Adaptive Cruise Control (Highway) & 0 & 100 & 11.06 \\
 \hline
   Collision Detection (Urban) & 0 & 30 & 27.66 \\
 \hline
\end{tabular}
\caption{\label{tab:metric-zones} Automotive Metric Zones}
\end{table}
\end{center}

\section{Results}\label{sec:results}

\subsection{Quantitative Results} 

\begin{table}[h]
	\centering
	\footnotesize
	\begin{tabular}{|p{1cm}|p{1cm}|p{1cm}|p{1.3cm}|p{1.2cm}|p{.7cm}|p{.8cm}|p{1cm}|p{1cm}|p{1.1cm}|}
		\hline
		Dataset & CNN & Model Input &  absErrorRel & sqErrorRel & iRMSE & SILog & Return Classifier Error & GFLOP & Parameter Size (MB) \\ [0.5ex]
		\hline\hline
		DSDepth & DSCnet & Mono camera & 5.77 & 4.16 & 8.04 & 11.22 & 4.79 & 8.79 & 82.21 \\
		\hline
		DSDepth & DSCnet & Stereo camera & 4.69 & 2.91 & 6.90 & 9.21 & \textbf{4.54} & 11.26 & 82.36 \\ 
		\hline
		DSDepth & DSCnet & Stereo + Scala & \textbf{4.37} & \textbf{2.77} & \textbf{6.86} & \textbf{8.89} & 4.61 & 12.62 & 85.24 \\ 
		\hline
	\end{tabular}
	\caption{\label{tab:sensor-set-comparison} DSCnet results with different sets of input sensors}
\end{table}

In Table \ref{tab:sensor-set-comparison}, we show results from various input sensor configurations across our evaluation metrics on the DSDepth test set.
As can be seen, with only a monocular camera as input, DSCnet achieves under 6\% absolute relative error.
Also, with each additional input, our model improves across all evaluation metrics other than the return classifier error.
This result is particularly exciting because of the minimal amount of effort required to incorporate new sensors.

\subsubsection{Automotive Metric Zones}

\begin{table}[h]
  \centering
  \footnotesize
 \begin{tabular}{|p{1cm}|p{1cm}|p{2cm}|p{1cm}|p{1.5cm}|p{1.5cm}|p{1.5cm}|}
 \hline
   Dataset & CNN & Model Input &  Parking Assist & Collision Detection & Adaptive Cruise Control & Overall \\ [0.5ex]
 \hline\hline
   DSDepth & DSCnet & Mono camera & 3.49 & 4.47 & 5.29  & 5.77 \\
 \hline
   DSDepth & DSCnet & Stereo camera & 3.00 & 3.61 & 4.30 & 4.69 \\ 
 \hline
   DSDepth & DSCnet & Stereo + Scala & 2.99 & 3.52 & 4.13 & 4.37 \\ 
 \hline
\end{tabular}
  \caption{\label{tab:automotive-zone-comparison} Relative Error (absErrorRel) for DSCnet in the automotive metric zones}
\end{table}


As discussed in Section \ref{sec:metric-zones}, we also evaluated our models using metrics designed for automotive use cases in Table \ref{tab:automotive-zone-comparison}.
As can be seen, our models achieved significantly lower relative error in each of the automotive metric zones when compared error across the entire point cloud.
This result shows areas that are both far away and in the vehicle's periphery account for the majority of the error.

\subsubsection{DSCnet-lite}

\begin{table}[h]
	\centering
	\footnotesize
	\begin{tabular}{|p{1cm}|p{1cm}|p{1cm}|p{1.3cm}|p{1.2cm}|p{.7cm}|p{.8cm}|p{1cm}|p{1cm}|p{1.1cm}|}
		\hline
		Dataset & CNN & Model Input &  absErrorRel & sqErrorRel & iRMSE & SILog & Return Classifier Error & GFLOP & Parameter Size (MB) \\ [0.5ex]
		\hline\hline
		DSDepth & DSCnet & Stereo Camera & 4.69 & 2.91 & 6.90 & 9.21 & 4.54 & 11.26 & 82.36 \\ 
		\hline
		DSDepth & DSCnet-lite & Stereo Camera & 6.42 & 6.51 & 11.31 & 14.44 & 5.50 & \textbf{2.30} & \textbf{1.83} \\ 
		\hline
	\end{tabular}
	\caption{\label{tab:embedded} Evaluation of our DSCnet-lite model for embedded devices}
\end{table}

As we mentioned earlier in the paper, CNNs have yielded a dramatic improvement of the state-of-the-art error-rate on a variety of computer vision tasks including image classification, object detection, semantic segmentation, and depth estimation.
However, CNNs often require far more resources (e.g. computation, memory, time, and energy) than previous computer vision methods.
This is of particular concern when deploying CNNs on embedded platforms such as smartphones, security cameras, and low-cost automotive-grade processors.
To mitigate this issue, researchers have developed resource-efficient CNNs such as SqueezeNet~\cite{SqueezeNet} and MobileNet~\cite{19} for image classification; YOLO~\cite{YOLO} and SqueezeDet~\cite{SqueezeDet} for object detection; and ENet~\cite{ENet} and SqueezeNet-based models~\cite{FelixNet} for semantic segmentation.
But, resource-efficient CNNs for depth estimation is a relatively untapped field.
To begin to address this opportunity, we have created a resource-efficient version of DSCnet called DSCnet-lite.

For brevity, we omit the precise dimensions of DSCnet-lite.
But, in order to reduce the number of parameters and floating point operations, one of the techniques behind DSCnet-lite is to replace dense convolutions with depthwise separable convolutions, similar to \cite{19}.

We show a quantitative evaluation of DSCnet-lite in Table~\ref{tab:embedded}. 
Going from DSCnet to DSCnet-lite, we have reduced the computational cost by 4.9x (to 2.3 GFLOP per inference), and we have reduced the quantity of parameters by 45x (to 1.83 MB).
This yields a modest increase in the error-rate (from 4.6\% absolute relative error for DSCnet to 6.4\% absolute relative error in DSCnet-lite). 
With our own CNN framework running on a garden-variety 4-core ARM A72 processor (found in millions of smartphones today) and without using any type of GPU or accelerator, we can routinely run CNN inference at 12.5 GFLOP/s, which implies that we should be able to run DSCnet-lite at over 5 inferences-per-second\footnote{We say inferences-per-second instead of frames-per-second, because we are talking about two-camera input in this example.} on a generic smartphone processor.
Further, many of today's server GPUs are able to run CNNs at much more than 1 TFLOP/s, but if we conservatively envision the case of running on a GPU at 1 TFLOP/s, we could do over 400 inferences-per-second with DSCnet-lite.


\subsection{Qualitative Results}

\begin{center}
  \begin{figure}
\newcolumntype{C}{>{\centering\arraybackslash}m{12em}}
  \begin{tabular}{l*4{C}@{}}
\toprule
    Example & DSCnet Output & HDL-64 Output & Input Image \\ 
\midrule
1 &\includegraphics[width=12em]{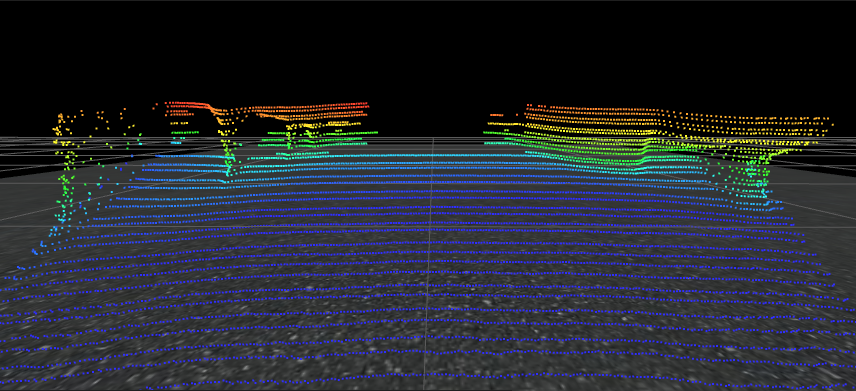} & \includegraphics[width=12em]{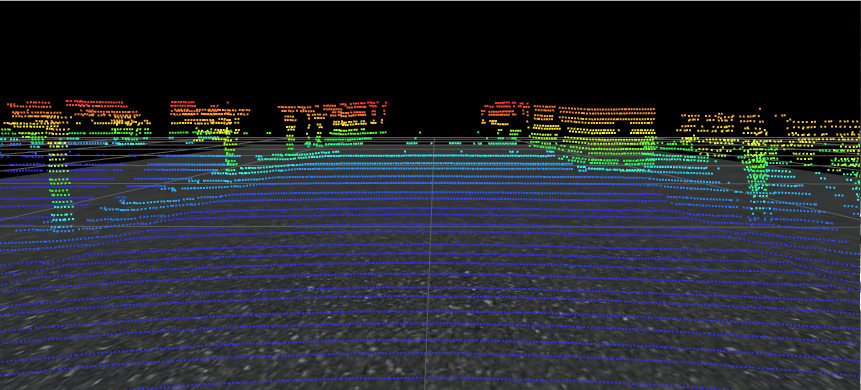} & \includegraphics[width=12em]{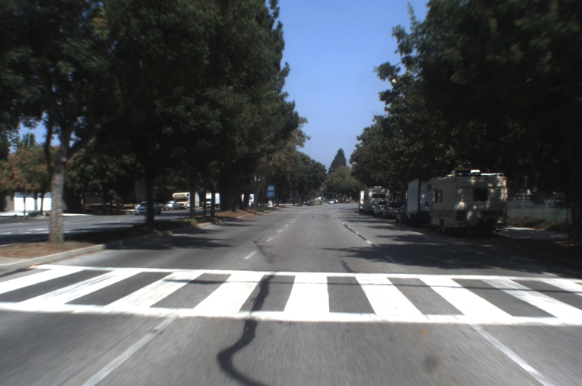} \\ 
2 &\includegraphics[width=12em]{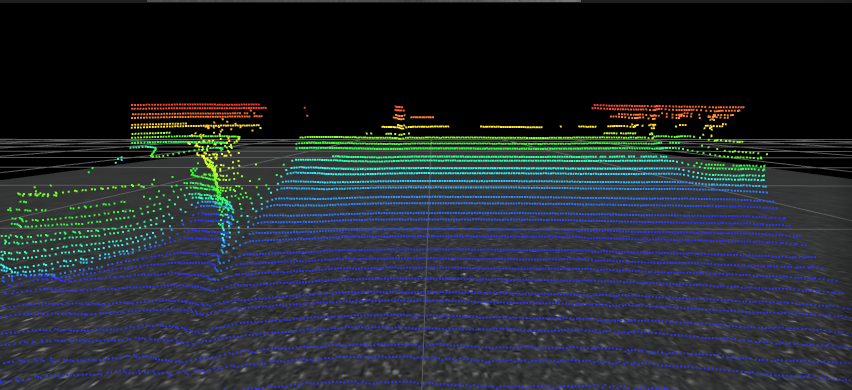} & \includegraphics[width=12em]{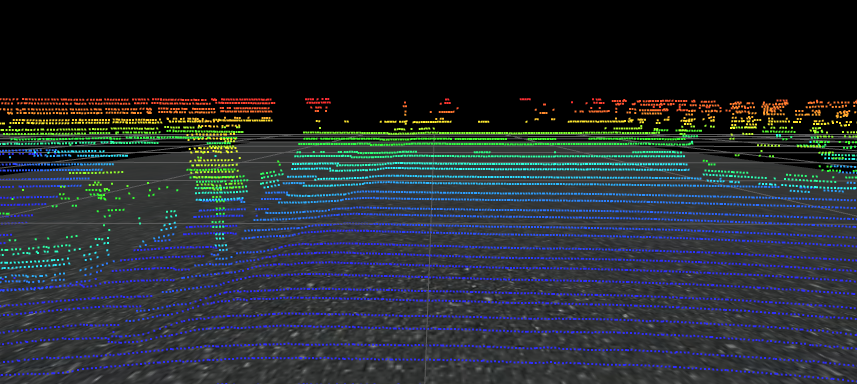} & \includegraphics[width=12em]{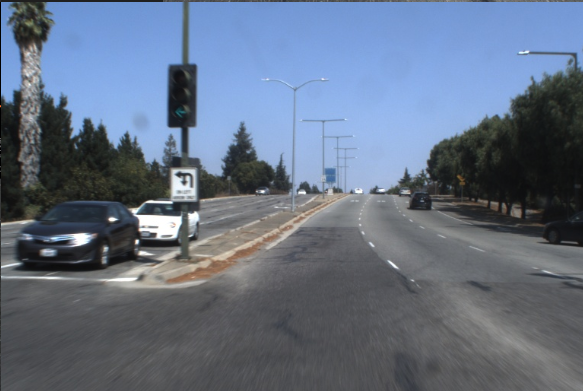} \\ 
3 &\includegraphics[width=12em]{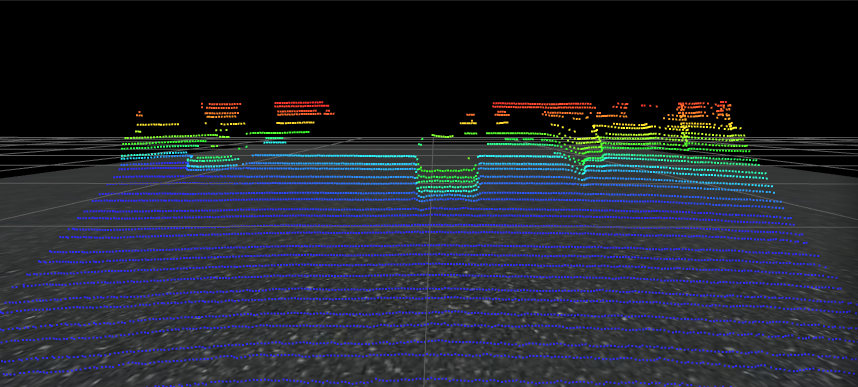} & \includegraphics[width=12em]{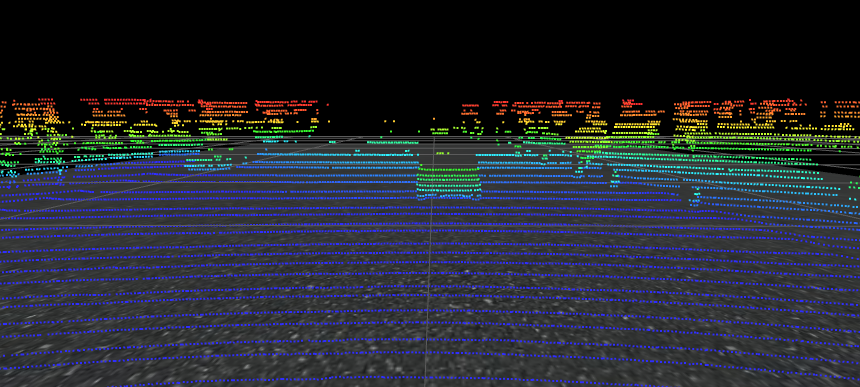} & \includegraphics[width=12em]{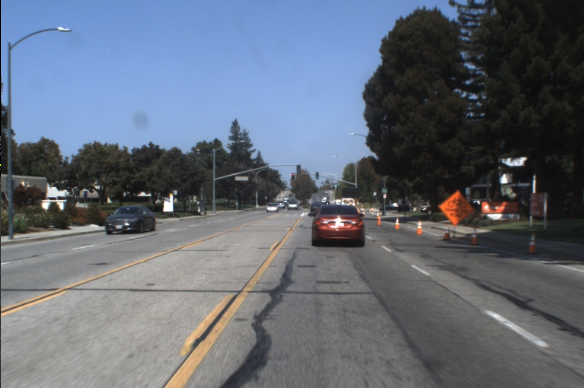} \\ 
4 &\includegraphics[width=12em]{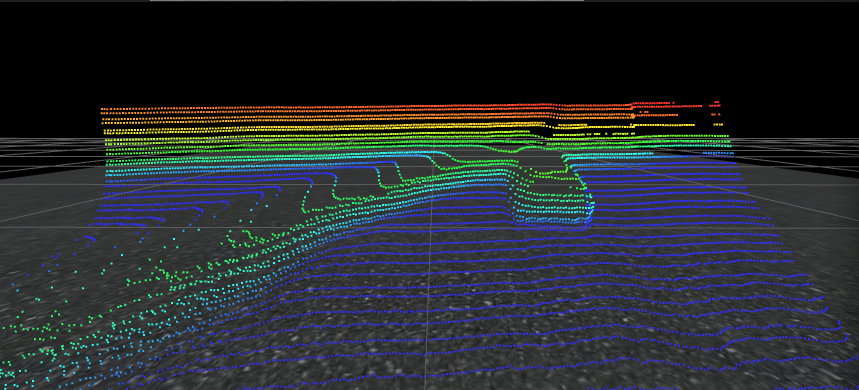} & \includegraphics[width=12em]{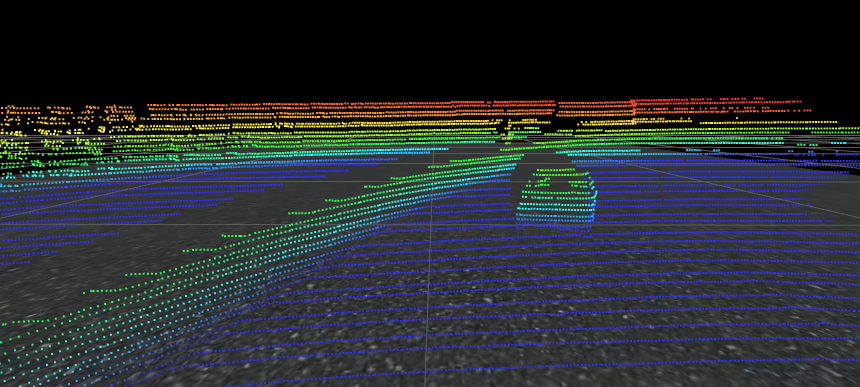} & \includegraphics[width=12em]{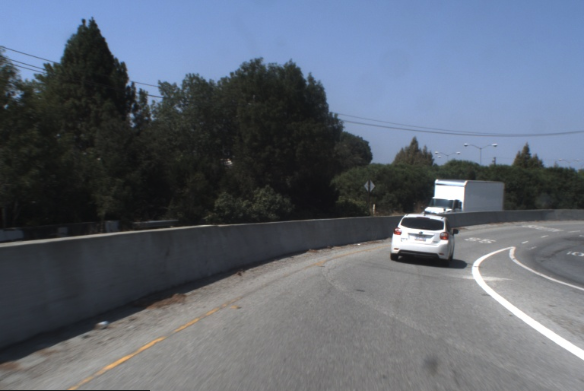} \\ 
5 &\includegraphics[width=12em]{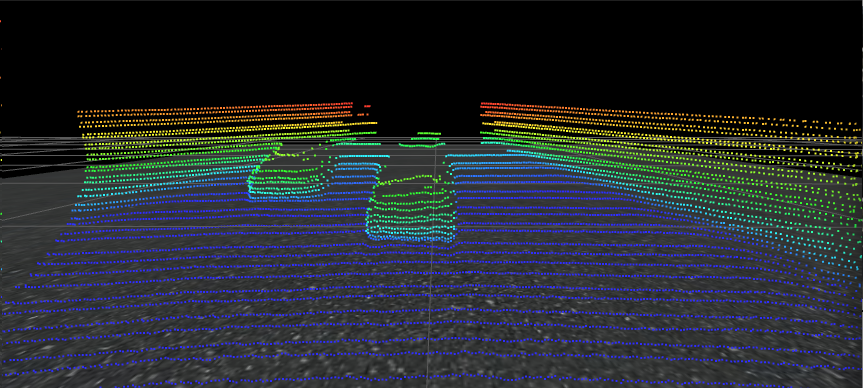} & \includegraphics[width=12em]{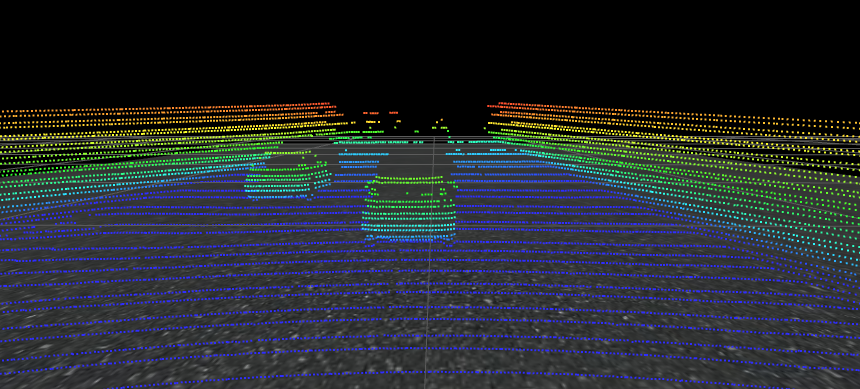} & \includegraphics[width=12em]{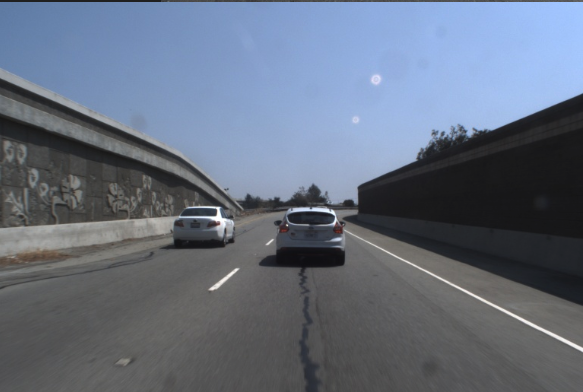} \\ 
\bottomrule 
\end{tabular}
  \caption{\label{fig:dsf-examples} Qualitative examples of DSCnet results compared to ground-truth data from a Velodyne HDL-64 LIDAR. In these results, the inputs to DSCnet are stereo cameras and Scala data.}
    \end{figure}
\end{center}

In Figure \ref{fig:dsf-examples}, we visualize the generated point-cloud from our of our DSCnet model (using two cameras and Scala data as inputs to DSCnet).

In Examples 1 and 2, you can see that DSCnet's generated point cloud looks similar to the ground truth LIDAR, and the model is able to distinguish the depths of objects such as cars, trees and traffic light poles and signs, as well as the ground plane. 

Example 3 shows our depth prediction near the beginning of a construction site. While our DSCnet model performs well on both the cars and the ground plane, DSCnet does not correctly predict the depth of the orange traffic cones along the right side of the road. 

In Examples 4 and 5, we visualize DSCnet results for predicting depth on the highway. In both examples, DSCnet is able to perceive the rough depth for the cars in front of the ego vehicle, as well as the road boundaries.

\section{Conclusion} \label{sec:conclusion}

Expensive sensors such as Velodyne HDL-64 LIDAR are commonly used in autonomous vehicle research.
However, due principally to their high cost, these expensive LIDARs are difficult to deploy in mass-market vehicles that are manufactured in the millions of units per year.
In this work, we have created a family of neural network architectures called DSCnet, which can be trained to "clone" expensive LIDAR while using only low-cost sensors as input.
We defined new metric zones for calculating distance predictions for the use of autonomous driving, and showed our DSCnet models could help perform certain perception tasks at a fraction of the price.
While LIDAR may still be needed for fully-autonomous driving, we feel that DSCnets running on low-cost sensors can provide high-quality real-time 3D data for semi-automation, or as a backup solution to systems relying on LIDAR.
Finally, we are interested to see how the emerging research field of Deep Sensor Cloning will impact the cost, quality, and reliability of autonomous vehicles and other applications.

\newpage

\small

\bibliographystyle{apa}
\bibliography{dsc}
\end{document}